\documentclass[runningheads]{llncs}

\usepackage{eccv}

\usepackage{eccvabbrv}

\usepackage{graphicx}
\usepackage{booktabs}
\usepackage{amsmath}
\usepackage{amssymb}
\usepackage{pifont}
\usepackage{tabularx}

\usepackage{array}
\usepackage{verbatim}
\usepackage{tikz}
\usepackage{multirow}
\usepackage{color}

\usepackage[accsupp]{axessibility}  

\usepackage[pagebackref,breaklinks,colorlinks,citecolor=eccvblue]{hyperref}

\usepackage{orcidlink}

\begin{document}

\title{Pre-Trained Model Recommendation for Downstream Fine-tuning} 

\titlerunning{Abbreviated paper title}

\author{Jiameng Bai\inst{1} \and
Sai Wu\inst{1}\and
Jie Song\inst{1} \and
Junbo Zhao\inst{1} \and
Gang Chen\inst{1} 
}

\authorrunning{J. Bai et al.}

\institute{ Zhejiang University\\
\email{\{baijiameng,wusai,sjie,j.zhao,cg\}@zju.edu.cn}}

\maketitle



\begin{abstract}
As a fundamental problem in transfer learning, model selection aims to rank off-the-shelf pre-trained models and select the most suitable one for the new target task. 
Existing model selection techniques are often constrained in their scope and tend to overlook the nuanced relationships between models and tasks. 
In this paper, we present a pragmatic framework \textbf{Fennec}, delving into a diverse, large-scale model repository while meticulously considering the intricate connections between tasks and models. The key insight is to map all models and historical tasks into a transfer-related subspace, where the distance between model vectors and task vectors represents the magnitude of transferability. 
A large vision model, as a proxy, infers a new task's representation in the transfer space, thereby circumventing the computational burden of extensive forward passes and reliance on labels. 
We also investigate the impact of the inherent inductive bias of models on transfer results and propose a novel method called \textbf{archi2vec} to encode the intricate structures of models.  
The transfer score is computed through straightforward vector arithmetic with a time complexity of $\mathcal{O}(1)$. Finally, we make a substantial contribution to the field by releasing a comprehensive benchmark. We validate the effectiveness of Fennec through rigorous testing on two benchmarks.  The code will be publicly available in the near future.
\end{abstract}

\section{Introduction}
The rapid advancement in deep learning has led to the proliferation of neural models across the internet. These models boast intricate architectures and require substantial computing resources for training. Consequently, many AI practitioners opt to utilize these pre-trained models for fine-tuning downstream tasks instead of embarking on training from scratch. This approach has become the de-facto paradigm in transfer learning: initiating with pre-training the foundational model on extensive datasets such as ImageNet~\cite{deng2009imagenet} and CIFAR10~\cite{krizhevsky2009cifar10}, followed by fine-tuning the pre-trained model for specific target tasks. Transfer learning has not only demonstrated enhanced performance~\cite{vasquez2021transfer} but also significantly accelerates the training process~\cite{raghu2019transfusion}.

The crux of the transfer learning scenario lies in selecting an appropriate pre-trained model for the target task.
This decision directly impacts the performance of the downstream task; an improper choice may even lead to negative transfer issues~\cite{Wang_2019_CVPR_negtive}. Undoubtedly, the most accurate method for model selection involves fine-tuning each model for the specific target task. 
However, given the extensive model repositories such as Huggingface\footnotemark[1] and Imgclsmob\footnotemark[2], this approach becomes computationally intensive and impractical. 


In this domain, there exist two primary approaches: computation-intensive and computation-efficient methods. The former, exemplified by taskonomy~\cite{zamir2018taskonomy}, aligns with the conventional definition of transfer learning. While subsequent studies~\cite{achille2019task2vec,song2019deep,rsa@2019,dds@2020} reduce computational requirements to some extent, they still involve optimization or fine-tuning, practices we aim to avoid.
In contrast, a pioneering work of the computation-efficient category is NCE~\cite{tran2019transferability_nce}, which employed conditional entropy between two task-related variables as the transferability score. Subsequently, researchers~\cite{nguyen2020leep_leep,you2021logme,bao2019information_hscore,huang2022frustratingly_transrate,pandy2022transferability_gbc,bolya2021scalable_parc, kim2016learning, gholami2023etran, li2023exploring_PED, wang2023far_ncti, zhang2023model_spider} explored diverse metrics to deduce this score. Most of them assess the correlation between forward features (generated by the target dataset and candidate models) and labels, employing various principles from information theory. For a detailed overview of related work, please refer to Appendix \ref{related}. Despite the advancements these methods have made, we identify three limitations in prior works:
\footnotetext[1]{\url{https://huggingface.co/}}
\footnotetext[2]{\url{https://github.com/osmr/imgclsmob}}
\begin{itemize}
    \item \textbf{Limitation 1: Non-transferable Score}. Most of the aforementioned methods fall into this category, signifying that the transferability estimation made for the current task holds no relevance for both previous and subsequent tasks. However, studies such as~\cite{alvarez2020geometric_ot,nguyen2020wide} have shown that models perform similarly in closely related domains. Additionally, models sharing similar architectures, like the ResNet family~\cite{he2016resnet}, exhibit comparable inductive biases, resulting in similar transferability. Broadly speaking, these methods tend to overlook the underlying relationships among models and historical tasks.
    \item \textbf{Limitation 2: Lack of Labels}. As most of the aforementioned methods rely on the correlation between forward features and labels, they encounter challenges when dealing with limited label data, a common scenario in real-world applications. Only few works \cite{gholami2023etran} can handle unlabelled scenario. Moreover, when dealing with extensive candidate models, conducting forward propagation for all models can result in substantial time overhead. 
    \item \textbf{Limitation 3: Limited Search Scope}. The model selection space of existing works is limited, and it is unknown how they will perform on a large model repository. To achieve a comprehensive ranking result, we need to evaluate models in a more diversified model space. 
\end{itemize}

To overcome these limitations, we propose a novel and efficient model ranking framework \textbf{Fennec}, comprising the Transfer, Meta, and Merge phase.

\textbf{Transfer phase.} Analogous to \emph{recommendation systems} \cite{sarwar2001item_cf}, our key insight is to construct transferability-related subspace from historical performances \emph{(interaction)} data. We maintain a historical performance matrix, where models and tasks are mapped into the latent subspace obtained through matrix decomposition. To mitigate the laborious fine-tuning process, we employ Fisher Discriminant Analysis to derive performances. Within this subspace, the efficient inner product of tasks \emph{(users)} and models \emph{(items)} vectors serves as the transfer scores \emph{(rating scores)}. \textbf{[Limitation 1]} 

\textbf{Meta phase.} We recognize the significant impact of a model's inherent expressive ability on its transferability. We propose the archi2vec method. It applies a directed acyclic attribute graph to represent the model's complicated structure. We employ the graph encoding method to convert the structures of models into vectors. We then extract meta-features from models by incorporating some accessible statistical features along with architectural features learned by the archi2vec method. \textbf{[Limitation 1]}

\textbf{Merge phase.} When conducting transferability evaluations on the target task \emph{(cold start)}, we align the new task into the transfer latent space. To adapt to the unlabeled scenario and avoid the time-consuming forward passes on all candidate models for a new task, we use a large vision model as a proxy to infer the vector of the new task in the transfer space. Even when solely utilizing the scores from the transfer phase, our method achieves comparable results to the majority of baselines. By merging the features of the meta phase, our approach achieves the state-of-the-art results with $\mathcal{O}(1)$ time complexity. \textbf{[Limitation 2]} 

\textbf{Benchmark.} We establish an extensive benchmark that includes 105 pre-trained models, spanning over 60 different architectures. The effectiveness of the framework has been validated on both public benchmark and our large-scale benchmark. We release it to foster advancements in this field. \textbf{[Limitation 3]}

Our contributions are summarized as below:
\begin{itemize}
    \item We propose a novel efficient framework for pre-trained model ranking. We draw inspiration from the recommendation systems, inferring latent vectors for both models and tasks in the transfer space. The inner product between the model and task vectors serves as the transfer scores. This process eliminates the need for forward features and labels on the new task, placing this work in the domain of unsupervised model ranking—a less-explored avenue.
    \item We propose the archi2vec method to automatically encode the complex architecture of neural networks into vectors, confirming the impact of the inherent inductive bias of models on model transferability.
    \item We establish a new benchmark, encompassing the largest model selection space to date, and make this benchmark publicly available. Our framework achieves state-of-the-art results on two benchmarks with minimal time cost.
\end{itemize}

\begin{figure*}[tbp]
	\centering
        \includegraphics[width=1\textwidth]{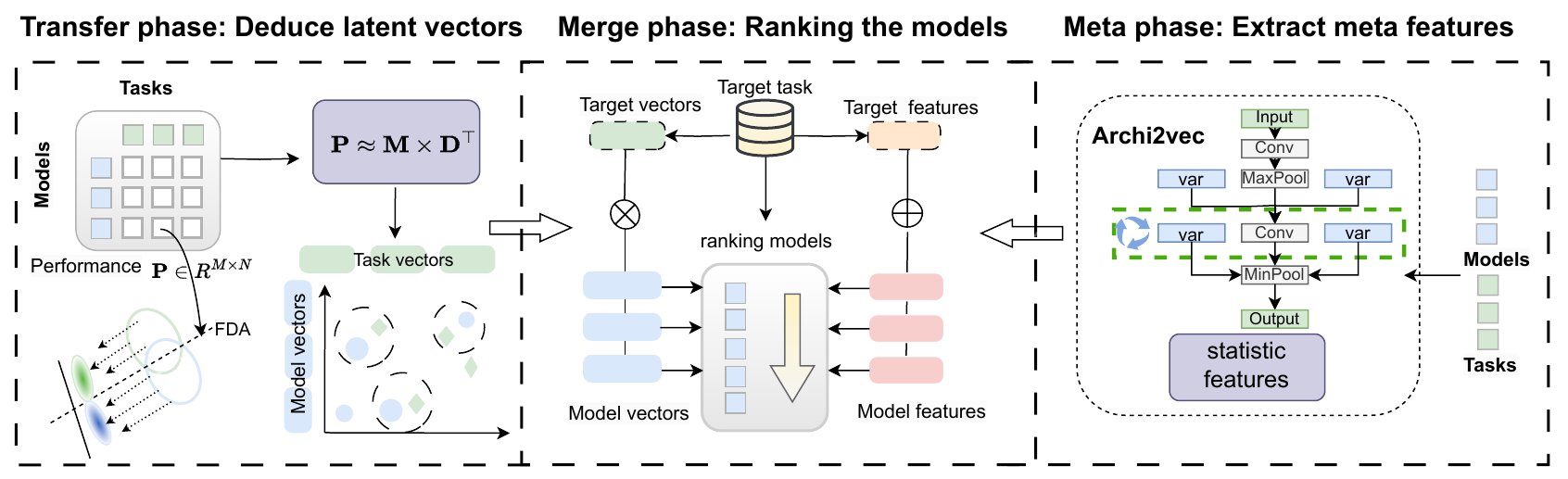}
	\caption{The overview of Fennec framework. We simultaneously consider the impact of forward features (Transfer phase) and intrinsic properties (Meta phase) on transferability estimation while conducting model ranking with high efficiency (Merge phase). }
	\label{fig:framework}
 \vspace{-1em}  
\end{figure*}

\section{Preliminaries}

\textbf{Notations.}  
The extensive model repository, denoted as $\mathcal{M}=\{\phi_i = (\theta_i \circ h_i)\}_{i=1}^M$, comprises $M$ pre-trained models. Each model encompasses a source task-related classification head $h_i$ and a feature extractor $\theta_i$ that generates $D$-dimensional features. The training data for these pre-trained models is inaccessible. On the other hand, the target dataset is represented as $\mathcal{T}_t=\{(x_j, y_j)\}_{j=1}^L$. Our goal is to identify the optimal feature extractor $\theta_i$ to align with the target dataset, thereby transferring the knowledge from the source model to the task domain. During the fine-tuning phase, the new feature extractor, denoted as $\Tilde{\theta}$, is initialized with the selected model $\theta_i$. Additionally, a new task-specific head, $\Tilde{h}$, is retrained on the target task alongside $\Tilde{\theta}$.
Given the pre-trained model, the optimal model is:
\begin{equation}
     \theta_i^* \circ h_i^* = \underset{\Tilde{\theta} \circ \Tilde{h} \in \mathcal{W}}{{\arg\max} } \, \frac{1}{n} \sum_{j=1}^L \log p(y_j|x_j;\Tilde{\theta},\Tilde{h},\theta_i) .
\end{equation}
where $\mathcal{W}$ is the parameters space.

\vspace{0.5em} 
\noindent\textbf{Problem Setting.} 
The transferability of ${\theta_i}$ to the target task is represented by the expected accuracy of $\theta_i^* \circ h_i^*$ on the test set of the target dataset, denoted as $\{a_i^t \in {a^t}\}_{i=1}^M$. Our objective is to estimate this transferability score to rank all the $\theta_i$ in the entire model repository without any fine-tuning operations, indicated as $\{s_i^t \in {s^t}\}_{i=1}^M$. The final evaluation metric is the Pearson correlation coefficient ($\tau$) between the ground truth ranking $a^t$ and the estimated ranking $s^t$.

\vspace{0.5em} 
\noindent\textbf{Methods Review.}   
We compare the paradigms of existing works with ours. To estimate the transfer scores of all the models on the target task $\mathcal{T}_t$, existing methods firstly need to obtain the forward features of each model on the target task $\{f_{it}=\theta_i(x_j \in \mathcal{T}_t)\}_{i=1}^M$. The estimated transfer score is evaluated by calculating sophisticated information-theoretic metrics between the forward features of an individual model and the labels. For instance, TransRate~\cite{huang2022frustratingly_transrate} employs mutual information for correlation metric. When computing with the full dataset, a total of $M*L$ forward passes are required. Our method estimates scores at a high-level semantic space which eliminate the need for forward pass on all model-task pair.

\section{Method}
In this section, we propose a novel model ranking framework consisting of three phases: Transfer phase, Meta phase, and Merge phase, as shown in Figure \ref{fig:framework}. 
In transfer phase (offline), we derive the latent vectors of models from the decomposed transfer space. In the meta phase (offline), we explore the intrinsic meta-features of models that influence transferability and propose the archi2vec method to encode the complex architectures of models. In the merge phase(online), we access the transferability of models to new task. To eliminate the dependence on forward features and labels, we adopt the large model as a proxy to infer the vectors of new task, while considering the impact of meta features.

\subsection{Transfer Phase}\label{sec:transfer_phase}
Transferring a model to new task is influenced by various intricate factors, including domain similarity and the training process. We treat these factors as learnable latent variables, extracting transfer preferences from historical data.

\vspace{0.5em}
\noindent\textbf{The historical performance matrix.} 
We draw inspiration from recommendation systems, where the key lies in unveiling latent vectors of users (tasks) and items (models) from historical interaction data. These vectors encapsulate user and item preferences, and their efficient combination serves as a measure of compatibility, facilitating the recommendation of new items to users. In this phase, our primary aim is to uncover these latent preferences related to transfer-influencing factors. Therefore, our first goal is to construct a historical performance matrix to deduce these transfer preferences.

We maintain the $N$ historical tasks, denoted as  $(\hat{\mathcal{T}_j}=\{(\hat{x_a}, \hat{y_a})\}_{a=1}^{n_j})_{j=1}^N$. The forward features extracted by $\theta_i$ are denoted as $f_{ia}=\theta_i(\hat{x_a})$. The transfer performance of all models in the model repository $\mathcal{M}=\{(\theta_i \circ h_i)\}_{i=1}^M$ across historical tasks constitutes the historical performance matrix $\mathbf{P}\in \mathbb{R}^{M\times N}$. Each item $p_{ij}$ in $\mathbf{P}$ denotes as transferability of model $i$ on the task $j$.
However, fine-tuning all models on tasks to obtain actual performances is resource-intensive. To mitigate this challenge, we employ Fisher Discriminant Analysis (FDA)~\cite{mika1999fisher}, as inspired by SFDA~\cite{shao2022not_sfda}, to estimate real transfer performance.
As illustrated in Figure \ref{fig:fda}, we visualized the classification performance of FDA. The results are consistent with the visualization of SFDA, demonstrating that FDA also exhibits good separability, rendering FDA a dependable proxy for actual performance. In fact, the values in the performance matrix can be the values of any existing evaluation methods, such as LogME, TransRate, etc.

Mathematically, FDA defines a projection matrix $U\in \mathbb{R}^{D\times D'}$ that projects origin data onto the subspace spanned by the column of $U$. 
The optimization of FDA can be solved efficiently by generalized eigenvalues problem~\cite{ghojogh2019eigenvalue, ghojogh2019fisher}.

Subsequently, we transform raw features using the learned projection matrix $U$ as follows: $\Tilde{f_{ia}}:=U^Tf_{ia}$. The performance of these transformed features can be obtained using Bayes' theorem~\cite{shao2022not_sfda}. Specifically, for each class $c$ in task $j$, $\mu_c$ represents the transformed mean value. We simplify $\Sigma$ to $I$ for clarity(as SFDA did). The probability that the data belongs to the class $c$ is then calculated as follows. Please refer to the Appendix \ref{sec:fda_deduce} for detailed derivation.
\begin{equation}\label{eq:fda_final}
    p(\hat{x_a}, c) = \Tilde{f_{ia}}^T\Sigma^{-1}\mu_c - \frac{1}{2}\mu_c^T\Sigma^{-1}\mu_c + log(\pi_c) ,\\
\end{equation}
where $\pi_c=\frac{n_c}{n_j}$ is the prior probability.  
Finally, the FDA score $p_{ij} $ can be obtained by the normalization and summation on the whole dataset:
\begin{equation}
        \label{eq:p_ij}
        p_{ij} = \sum_{a=1}^{n_j} \frac{{\rm exp}^{p(\hat{x_a}, \hat{y_a})}}{{\sum_c {\rm exp}^{p(\hat{x_a}, c)}}} .
\end{equation}

\vspace{0.5em}
\noindent\textbf{The hidden embeddings.} To obtain the latent transfer subspace, we decompose the historical FDA matrix to learn transfer embeddings for each pre-trained model $\boldsymbol{m_i}$ and historical dataset $\boldsymbol{t_j}$. Specifically, we use the Non-negative Matrix Factorization(NMF) \cite{lee2000algorithms_nmf} to decompose the matrix because all the elements in the historical performance matrix $\mathbf{P}$ are non-negative values (Equation\eqref{eq:p_ij}), which perfectly adheres to its requirement for non-negativity. The convergence of NMF has been proven in the works \cite{lee2000algorithms_nmf, fevotte2011algorithms_nmfcd}.
The objective function is:

\begin{equation}
     \mathop{\rm min}\limits_{\mathbf{M,D}} \quad || \mathbf{P} - \mathbf{MD}^\top ||_F^2 + \alpha_m ||\mathbf{M}||_F^2 + \alpha_d ||\mathbf{D}||_F^2 . \\
\label{eq:nmf}
\end{equation}
where $\mathbf{M} \in \mathbb{R}^{M\times k}$ and $\mathbf{D} \in \mathbb{R}^{N\times k}$ are the learnable model embedding matrix and the task embedding matrix, respectively. The dimension of latent embeddings is $k\in \mathbb{R}$, and we use $||.||_F^2$ to denote the Frobenius norm. 

In essence, we map the key factors influencing transferability into $k$-dimensional transfer space and represent all candidate models and historical tasks as vectors in this space. Here, $\boldsymbol{d_j}$ represents the preferences of task $j$ regarding these factors, while $\boldsymbol{m_i}$ signifies the extent to which these factors are met. A higher value of $\boldsymbol{m_i}\cdot \boldsymbol{d_j}$ indicates a stronger correlation between model $i$ and task $j$.

\subsection{Meta Phase}\label{sec:meta_phase}
\footnotetext[3]{https://github.com/sksq96/pytorch-summary }

 \begin{figure}[tbp]
	\centering
        \includegraphics[width=0.8\textwidth]{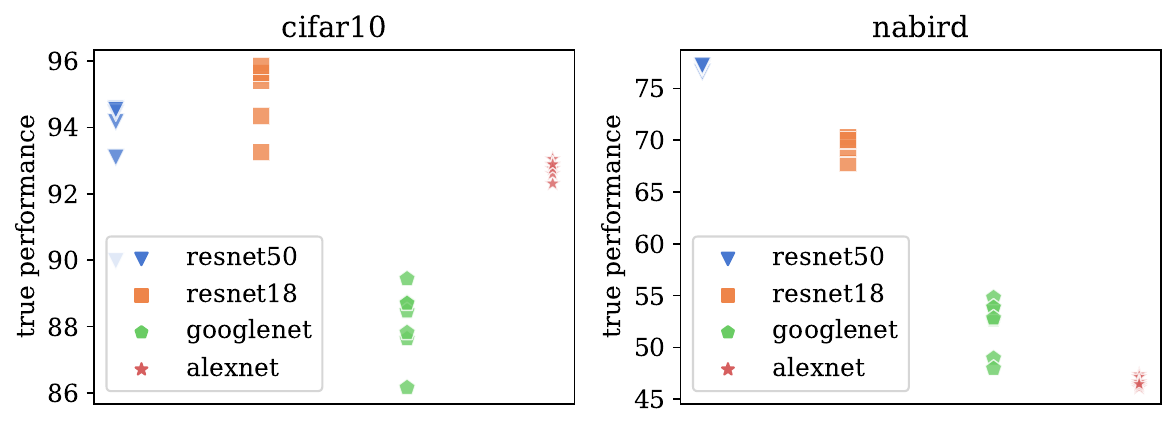}
	\caption{The real transfer effects of diverse model architectures on Cifar10 and Nabird. The structure of the model exhibits a discernible connection with its transfer result.}
	\label{fig:meta}
        \vspace{-1em}  
\end{figure}

The architecture is an inherent attribute of the model. From the Figure \ref{fig:meta},
we observe that architectures of neural models also affect their transfer performances on different tasks. we propose an architecture encoding method \textbf{archi2vec} to automatically extract high-level features of complex architecture information.

We introduce two types of meta-features: basic and advanced. In the basic type, for models, we consider parameter size and layer size; for tasks, we consider class number and data volume.
Advanced feature, on the other hand, represents the model structure. 
Deep learning models possess unique inductive biases due to their inherent architectural properties, significantly impacting model generalization~\cite{matsoukas2022bias}. 
Utilizing one-hot encoding to represent different architectures is inadequate for two reasons: (1) it becomes unwieldy as the model space expands; (2) it fails to capture the nuances between model architectures. To address this, we propose the novel archi2vec method to automatically infer model architecture information. This method comprises the following two key components.

\textbf{(1) The runtime graph of pre-trained models.} 
Neural networks exhibit diverse depths, involve numerous complex operations (Activation, BatchNorm, Pooling), and feature intricate connections (such as skip connections in the ResNet family~\cite{he2016resnet}), rendering the encoding of model architectures an exceedingly challenging task.
One straightforward approach to encode model architecture involves using the static structure ($\rm summary$\footnotemark[3]) and treating the model as a sequence structure, stacked with some basic modules. However, this method fails to capture the meaningful structure due to its inability to handle skip connections and the actual path of data flow. To address this limitation, we propose a graph structure to represent the model architectures.

\begin{figure}[tbp]
	\centering
        \includegraphics[width=0.7\textwidth]{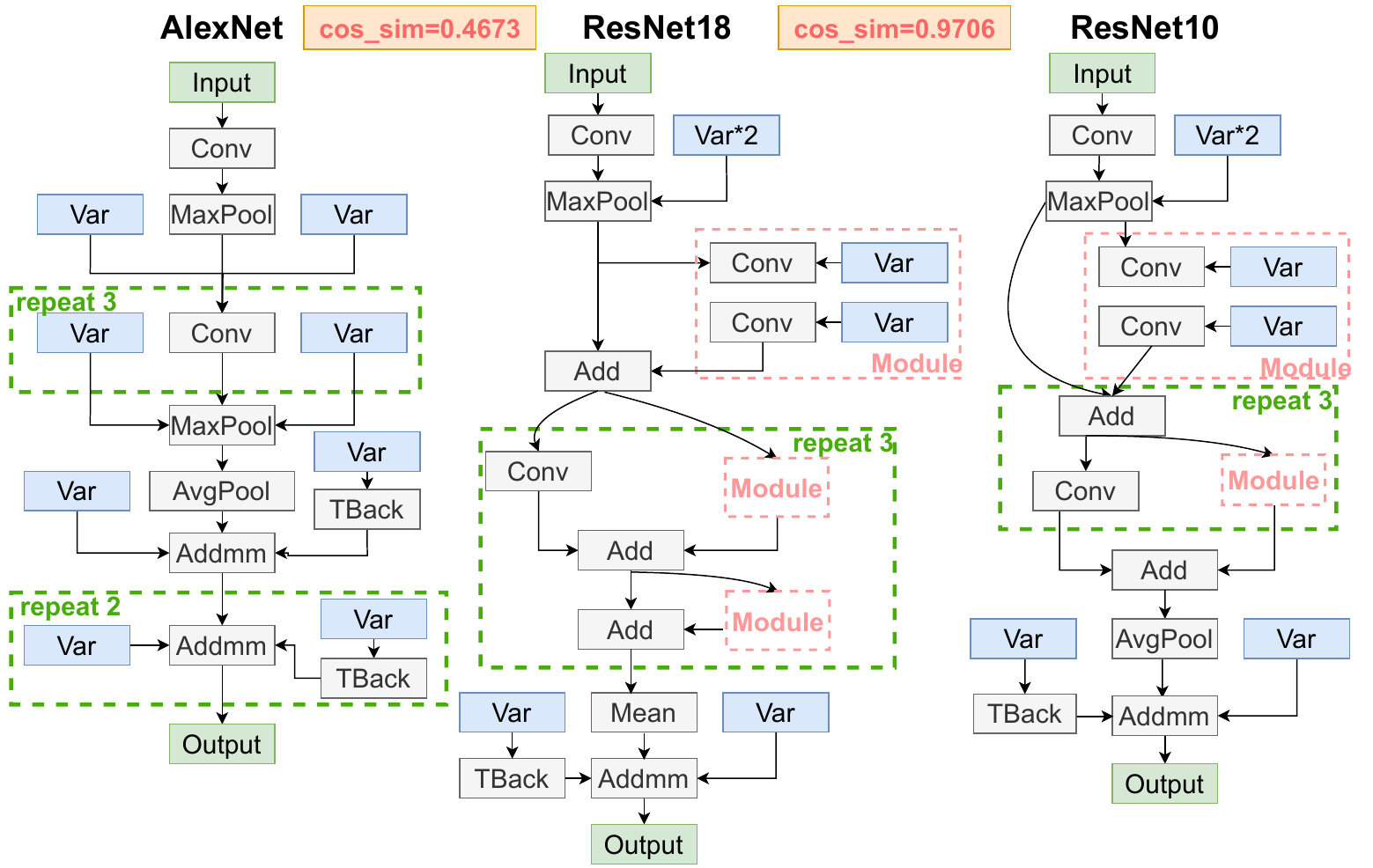}
	\caption{The directed acyclic attributed graph established for ResNet18, ResNet10 and AlexNet, omitting the BatchNorm, Activation, and other subtle operations for the sake of simplicity(they are present in the actual graph construction).}
	\label{fig:archi}
        \vspace{-1em}  
\end{figure}

We represent the architectural structure of models $\{\phi_i\}_{i=1}^M$ using a \textbf{directed acyclic attributed graph} denoted as $\{G_i=(V_i, E_i, A_i)\}_{i=1}^M$, where $V_i$ and $E_i$ represent the vertices and edges of the graph respectively, and $A_i$ signifies the attributes of nodes. Specifically, by feeding just one suitable input into models and leveraging the chain rule alongside modern deep learning frameworks' gradient propagation methods, we can discern the crucial operations (attributes) the data passes through (nodes) and the pathways along which data flows (edges). In practice, we recursively derive the entire structure from the output to the input. Figure \ref{fig:archi} shows the model architectures of AlexNet, ResNet18 and ResNet10.

We categorize node types as attributes. Based on our statistical analysis (Table \ref{tab:atoms}), all node types fall into a set of 37 modules: $(A_i)_{i=1}^M \in \mathcal{A}=(atom_i)_{i=1}^{37}$. It's important to note that we focus on high-level structures, omitting specific details such as the convolution kernel size (e.g., 3x3 or 1x1). 

\textbf{(2) Encoding the graph.} Our goal is to capture both the topological and semantic similarities between the two architectures, focusing on graph edges and associations between node attributes. To achieve this, we employ $\rm graph2vec$~\cite{narayanan2017graph2vec, karateclub}, a method analogous to $\rm doc2vec$~\cite{le2014doc2vec}. In this approach, the entire graph is treated as a document, and root subgraphs are treated as sentences, enabling unsupervised learning of the entire graph's embeddings. While $\rm graph2vec$ was initially designed for graph and node classification, we adapt it for our purposes by replacing the labels $\lambda$ in~\cite{narayanan2017graph2vec} with attributes $A_i$. The resulting graph embeddings are denoted as ($\boldsymbol{a_i})_{i=1}^M$. Finally, we utilize KMeans~\cite{krishna1999kmeans} to cluster these embeddings, effectively categorizing existing architectures. 

The advanced architectural features learned by archi2vec complement basic meta features, enhancing the representation of pre-trained models. Ultimately, in this phase, We concatenate the meta features of the pre-trained model as $\boldsymbol{f_i^{model}}$, and the meta features of historical task as $\boldsymbol{f_j^{data}}$.

\subsection{Merge Phase}
We formally evaluate the rankings of models when facing the target task $\mathcal{T}_t$.

\vspace{0.5em}
\noindent\textbf{Offline: Embedding mapping for tasks.} When dealing with a target task, akin to the recommendation system, we confront the pivotal \emph{'cold-start'} problem: determining the representation of a new task. Even within the realm of recommendations, this remains a challenging issue. To circumvent the dependency on task labels(which are typically costly to obtain), we propose a novel 'proxy estimation' method to map the target task into the $k$-dimensional transfer space. After mapping, the transfer preferences of target task can be evaluated within the same context as the entire set of pre-trained models.

Similar tasks inherently share comparable domain characteristics, and their latent transfer preferences should align accordingly. Measuring domain characteristics of tasks without auxiliary information poses a challenge. To address this, we employ one common proxy model to generate forward features for different tasks. The similarity in forward features produced by diverse tasks on the same model acts as a proxy, indicating the similarity between datasets. 

We aim for the proxy model to encompass a broad, diverse domain, ideally including domains similar to the target task, while minimizing additional training burdens. Therefore, we leverage large vision model such as CLIP (Contrastive Language-Image Pre-Training)~\cite{radford2021learning_clip, schuhmann2022laionb_openclip} as proxy model.
We sample a small subset $\hat{t}_j$ of historical task $\hat{\mathcal{T}}_j$.
The features of historical tasks extracted by CLIP are:

\begin{equation}\label{eq:clip}
    \boldsymbol{g_j} = {\rm CLIP.encode\_image}(\hat{t}_j) ,
\end{equation}

To establish the connection between the clip features and vectors in the transfer space, we construct a regressor that connects the CLIP features $\boldsymbol{g_j}$ with the learned embedding $\boldsymbol{d_j}$ in Equation\eqref{eq:nmf}. In practice, a random forest regressor is employed. The fundamental concept here is that akin tasks display similar features on proxy model, indicating analogous latent transfer preferences.  All historical tasks are utilized to train the regressor:
\begin{equation}\label{eq:regressor}
    \boldsymbol{d_j} = {\rm regressor}(\boldsymbol{g_j}) ,
\end{equation}

\vspace{0.5em}
\noindent\textbf{Online: Score of Transfer Phase.} When addressing the target task $\mathcal{T}_t$, we only need to obtain the CLIP features $\boldsymbol{g_t}$ using Equation\eqref{eq:clip} and deduce the task embedding $\boldsymbol{d_t}$ through Equation\eqref{eq:regressor}. Ultimately, the transfer scores can be directly computed by performing vector multiplication between the learned embeddings from the entire model repository and the new task embedding:

\begin{equation}
     \textsc{trans\_score}_{i,t} = \boldsymbol{m_i} \cdot \boldsymbol{d_t} .
\end{equation}

\vspace{0.5em}
\noindent\textbf{Offline: Mapping between meta features and transferability.}
To bridge the meta features and transferability, we train a Linear Regression (LR) model.  The transferability score $p_{ij}$ between model $i$ and task $j$ has inferred in Equation\eqref{eq:p_ij}. Meta features  $\boldsymbol{f_i^{model}},\boldsymbol{f_t^{data}}$ have defined in Section \ref{sec:meta_phase}.
\begin{equation}
    p_{ij} = {\rm LR}([\boldsymbol{f_i^{model}}, \boldsymbol{f_j^{data}}]) ,
\end{equation}

\vspace{0.5em}
\noindent\textbf{Online: Score of Meta Phase.} To obtain scores in the meta phase, we only need to acquire the meta features $\boldsymbol{f_t^{data}}$ of the target task. The meta score is defined in Equation\eqref{eq:meta_score}.
Finally, the ranking score of a model $i$ on the target task $t$ is obtained by merging the scores of two phases, defined in Equation\eqref{eq:final}, where $\alpha$ (default value is 0.5) tunes the weight of meta scores and transfer scores. 
\begin{equation}\label{eq:meta_score}
    \textsc{meta\_score}_{i,t} = {\rm LR}([\boldsymbol{f_i^{model}}, \boldsymbol{f_t^{data}}]) .
\end{equation}

\begin{equation}
   s_i^t  =  (1 - \alpha) * \textsc{trans\_score}_{i,t} + \alpha * \textsc{meta\_score}_{i,t} .
    \label{eq:final}
\end{equation}

It is crucial to emphasize that the transfer phase, meta phase, as well as the training of regressor and LR, are conducted offline on the historical task-model pairs. 
Our approach achieves a time complexity of $\mathcal{O}(1)$, as it circumvents the forward passes for all model-task pairs during the online ranking stage. Instead, it necessitates processing only for the current task to derive the score. Even more noteworthy is our ability to rank new tasks without relying on labels, marking our foray into the realm of unsupervised pre-trained model selection.

\section{Fennec Benchmark}
PARC~\cite{bolya2021scalable_parc} is the first official model selection benchmark with different source domains and architecture combinations. However, the public version of PARC contains 32 pre-trained models, encompassing 4 distinct architectures, thereby limiting the diversity of available models. Therefore, we create a larger benchmark. The models within the PARC originate from known training domains, whereas our extensive Fennec benchmark is of unknown provenance. 

\vspace{0.5em}
\noindent\textbf{Models and Datasets.} Most existing works typically comprise several well-known model types offered by the PyTorch framework. Instead, we are dedicated to constructing a benchmark with a greater quantity and wider variety of model types.
Our criteria for selecting models aims to cover a wide range of architectures and sizes. Thus, as illustrated in the Table \ref{tab:expri_setting}, our model repository spans from well-known architectures like ResNet\cite{he2016resnet} to more niche architectures like HRNet\cite{sun2019deep_hrnet}. For popular models, we have included various variants. To balance model diversity and training time overhead, we select 105 models, representing over 60 architectures. Additionally, Fennec includes tasks from diverse domains for evaluation, including Scenes, Medical scenarios, and Gesture.

%
For each task, we conduct a rigorous grid search over all 105 models to perform transfer learning, aiming to obtain the true performance of models on the task and evaluate transfer estimation methods. 
Please refer to the Appendix \ref{training} for a detailed training process. The training is performed using GPUs (NVIDIA RTX A5000), and by rough estimation, it consumes approximately 28000 GPU hours.
However, for feature extraction and transfer estimation, we conduct all computations on CPUs to ensure accessibility for practitioners without high-end computational resources, consistent with the PARC benchmark.

\vspace{0.5em}
\noindent\textbf{Baselines.} 
Due to the continuous development of this field, many advanced feature-based model selection methods have emerged. We extend the PARC by adding:
LogME~\cite{you2021logme}, TransRate~\cite{huang2022frustratingly_transrate},  GBC~\cite{pandy2022transferability_gbc}, SFDA~\cite{shao2022not_sfda}, ETran~\cite{gholami2023etran}, NCTI~\cite{wang2023far_ncti} and PED~\cite{li2023exploring_PED}. 
Unfortunately, as of now, the Moder Spider~\cite{zhang2023model_spider} has not made the necessary
 meta-training data for their models publicly available, making it impossible for us to compare with them.
Finally, we use the following baselines:
\begin{itemize}
    \item Taskonomy-based: This line of work requires additional necessary training on the target dataset. It is computationally intensive and doesn't align with our targets. We choose two prominent methods: RSA~\cite{rsa@2019} and DDS~\cite{dds@2020}. 
    
    \item Probability-based: NCE~\cite{tran2019transferability_nce} and LEEP~\cite{nguyen2020leep_leep} use the conditional probability between the target labels and source labels to evaluate the transferability. 
    
    \item Feature-based: The central concept is to utilize the correlation between the forward features of the penultimate layer (generated by the target task and models) and the corresponding labels as the key estimation metric. As these methods represent mainstream research, Fennec covers a broad spectrum of feature-based methods, including H-Score~\cite{bao2019information_hscore}, LogME~\cite{you2021logme}, TransRate~\cite{huang2022frustratingly_transrate}, PARC~\cite{bolya2021scalable_parc}, GBC~\cite{pandy2022transferability_gbc}, SFDA~\cite{shao2022not_sfda}, ETran~\cite{gholami2023etran}, NCTI~\cite{wang2023far_ncti} and PED~\cite{li2023exploring_PED}.
\end{itemize}

Our framework computes a historical performance matrix. In experiments, when one task serves as the target, the others are considered as historical tasks.

\begin{table*}[tbp]
	\caption{Overall results. For each evaluation, we sample five probe sets(
including 500 samples) using different seeds and then report the average results. The calculate time and pearson correlation coefficient(PC) are the average results of 840 transfers(PARC) and 3675 transfers(Fennec), respectively. The feature extractor time is the cumulative time required for feature extraction in all experiments.
The results demonstrate the efficiency of our framework with \textbf{minimal time consumption}. Meanwhile, we achieved the best performance \textbf{without relying on labels}.}
	\centering
  \resizebox{\textwidth}{!}{
	\begin{tabular}{l<{\centering}m{2cm}<{\centering} | m{2cm}<{\centering}m{2.3cm}<{\centering}m{2cm}<{\centering} | m{2cm}<{\centering}m{2cm}<{\centering}r<{\centering}}
		\toprule
            \multirow{4}{2cm}{\centering Methods} & \multirow{4}{2cm}{\centering No labels} & \multicolumn{3}{c|}{PARC Benchmark} & \multicolumn{3}{c}{Fennec Benchmark}\\
            \cline{3-8}
		  ~ &~ & Features extractor time\bf (s) & Calculate time\bf(ms)  & Mean PC(\%) & Features extractor time\bf(s)  & Calculate time\bf(ms) & Mean PC(\%)  \\
		\midrule
            RSA  {\textcolor{violet}{ (3+5 Hours)} } & \ding{56}    &   4137.4  &   $63.9 \pm 5.6$  & $64.88 \pm 0.21$ &    22015.9 &  $57.9 \pm 5.8$ & $83.31\pm 0.19$  \\
            DDS  {\textcolor{violet}{ (3+5 Hours)} } & \ding{56}    &   4137.4  &   $30.4 \pm3.8$  &  $62.83 \pm 0.33$  &   22015.9 & $26.6 \pm 3.5$  &  $85.55\pm 0.14$ \\
            \midrule
            NCE    & \ding{56}    & 0      &   $1.7 \pm 2.0$  & $ 2.08 \pm 0.67$   &  - &  - & - \\
            LEEP    & \ding{56}  & 0       &   $1.1 \pm 1.1$  & $ 10.82 \pm 0.13$  &  0 &  $2.0 \pm 1.0$ &  $74.05\pm 0.12$  \\
            \midrule
            H-Score  & \ding{56} &   4137.4  &   $12.6\pm 3.9$  &  $ 55.95 \pm 0.55$   &  22015.9  &  $9.1 \pm 3.3$ & $84.98\pm 0.10$\\
            LogME   &  \ding{56} &  4137.4  &   $18.0 \pm 5.5$  &  $  55.09 \pm 0.35$  & 22015.9  &  $10.5 \pm 4.1$ &  $82.79\pm 0.11$\\
            TransRate &  \ding{56} &  4137.4  &    $15.9 \pm 5.0$  & $ -6.45 \pm 0.68$  &  22015.9  &  $9.3 \pm 3.6$  & $86.36\pm 0.13$ \\
            PARC     &  \ding{56} &  4137.4 &    $38.9 \pm 4.7$  & $  59.31 \pm 0.68$  &  22015.9  &    $34.0 \pm 4.1$ & $83.28\pm 0.13$ \\
            GBC      &  \ding{56} &  4137.4 &   \textcolor{violet}{23342.7\tiny {$ \pm 25012.7$} }   & $ 46.07 \pm 0.83 $  & 22015.9  & \textcolor{violet}{2840.0\tiny {$ \pm 4377.1$} }
            & $76.63\pm 0.39$  \\
            SFDA    &  \ding{56} &  4137.4 &   $41.9  \pm 18.5$  & $ 67.76 \pm 1.24$  &  22015.9  &  $20.2 \pm 9.0$ & $85.41\pm 0.14 $ \\
            ETran    &  \ding{56} &  4137.4 &   $28.9 \pm 9.1$  & $  55.57\pm 1.05$  &  22015.9  &  $18.4 \pm 7.2$ & $85.17\pm 0.16 $ \\
            NCTI    &  \ding{56} &  4137.4 &   $44.7 \pm 18.2$  & $ 54.87\pm 0.67$  &  22015.9  &  $26.5 \pm 9.7$ & $70.79\pm0.60 $ \\
            PED    &  \ding{56} &  4137.4 &   \textcolor{violet}{9230.8\tiny {$ \pm 14313.1$} }  & $  54.46\pm0.40$  &  22015.9  & \textcolor{violet}{9682.7\tiny {$ \pm 19587.9$} }  & $85.99\pm 0.11 $ \\
            \midrule
            Ours            & \ding{52} & \bf 635.6 &    $  2.2 \pm 0.1 $    & $ \bf 68.12\pm0.57 $  & \bf 420.7 & $5.1 \pm 0.1$  & $\bf 87.79 \pm 0.29$ \\
            
		\bottomrule
	\end{tabular}  }
	\label{tab:main_result1}
   \vspace{-1.5em}  
\end{table*}

\section{Experiments}
In this section, we use PARC and Fennec benchmark to validate our framework.

\subsection{Experiment results}

\vspace{0.5em}
\noindent\textbf{Overall Comparison.} The comprehensive experimental results for both benchmarks are detailed in Table \ref{tab:main_result1}. Several key findings emerge from our analysis:

(1) Feature Extraction Time: Feature-based methods necessitate the extraction of forward features, constituting a significant portion of the evaluation process. Notably, this step is time-intensive. Our experiments underscore that the feature extraction time is the dominant factor in the evaluation process. Across both PARC and Fennec, the time for feature extraction escalates linearly with the expanding model repository, evident from the rise in time as the number of models increases from 32 to 105.
Nevertheless, our approach bypasses the need for computationally intensive forward feature extraction. Instead, it learns the latent transfer preferences between models and tasks in an offline process. During inference, it simply acquires the task's features from CLIP, a considerably more efficient operation than extracting forward features for every model-task pairing. In contrast to the tedious time consumed by forward feature extraction, computational times at the millisecond level can be deemed negligible.

(2) Probability-based methods offer fast computations but sub-optimal performances. NCE relies on the information of the model's source domain, making it unfit for the Fennec benchmark. RSA and DDS demand additional fine-tuning (3 hours on the PARC benchmark and 5 hours on the Fennec benchmark), reducing their practicality. Feature-based methods show promise, yet prolonged feature extraction hampers efficiency. Our approach achieves top-tier performance with minimal time, offering a superior solution for large-scale model ranking.

(3) The experimental results in the Table \ref{tab:ab_module} reveal that relying solely on scores of transfer phase for estimation is insufficient. Meta features must be used as supplementary information to the transfer score, leading to a better performance.

\begin{figure*}[tbp]
	\centering
        \includegraphics[width=1\textwidth]{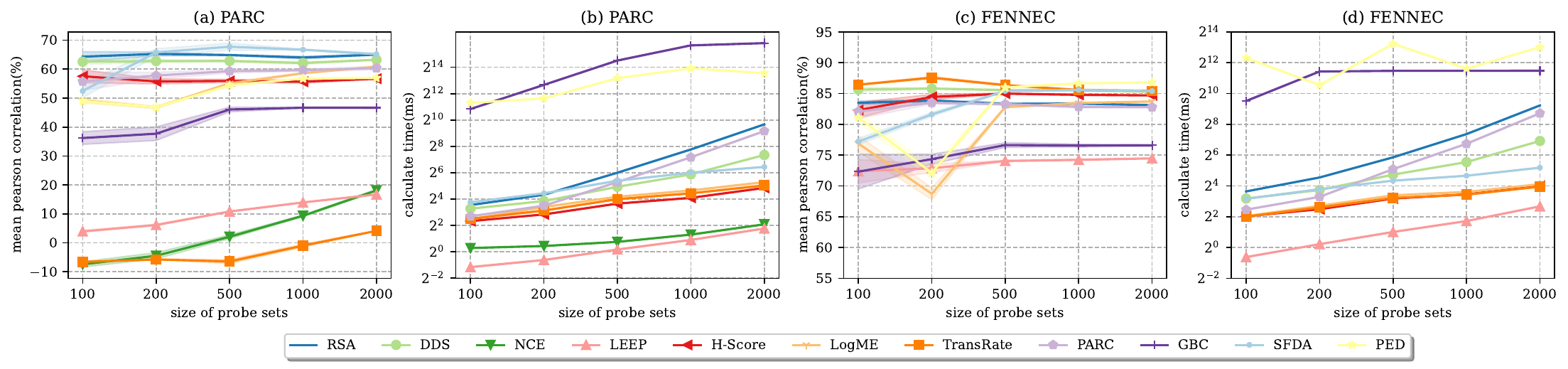}
	\caption{Comparison of performance and time varying on  probe sizes. As the probe dataset increases, the mean Pearson Correlation(PC) of most methods also increase. Figures(b)(d) reflect their slower computation time.}
	\label{fig:probeset}
 \vspace{-1em}  
\end{figure*}

\vspace{0.5em}
\noindent\textbf{Comparison of different probe sets.} 
Existing methods heavily rely on the correlation between forward features and labels. By default,  PARC and Fennec benchmarks use 500 samples as probe sets to generate forward features. To investigate the impact of the different sizes on the pearson  correlation and time consumption, we vary the size of probe sets from 100 to 2000.

\begin{figure}[tbp]
	\centering
        \includegraphics[width=1\textwidth]{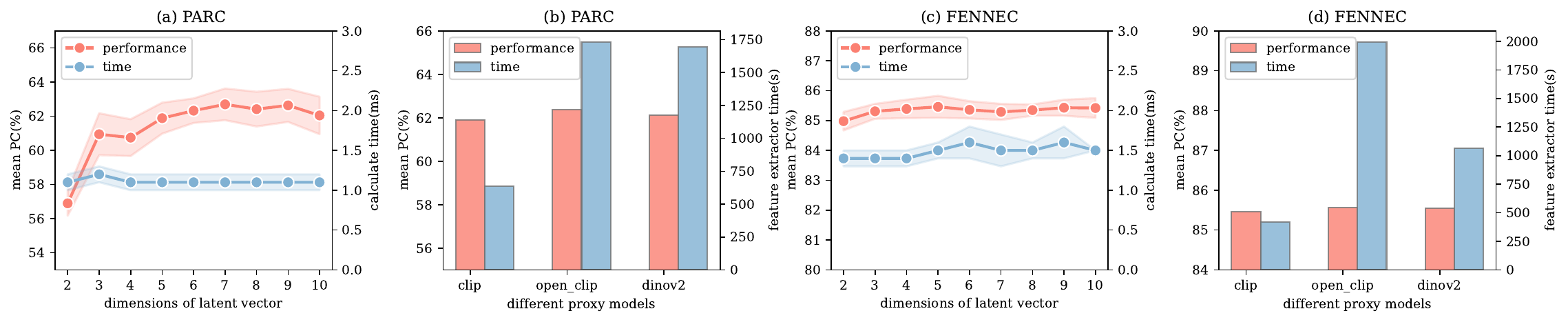}
	\caption{Figure(a)(c) reflects the influence of the dimension of matrix factorization on the mean PC. Figure(b)(d) reflects the impact of using different proxy models. }
	\label{fig:ks}
 \vspace{-1em}  
\end{figure}

The results for the PARC and Fennec benchmarks are shown in Figure \ref{fig:probeset}(a)(b) and Figure \ref{fig:probeset}(c)(d), respectively. As the probe set size increases, methods like NCE, LEEP, Transrate, GBC, and SFDA show improvements, underscoring the importance of probe set size as a limiting factor. These enhancements can be attributed to the improved fitting of the correlation function between features and labels with more samples. In contrast, RSA, DDS, and PARC methods maintain relatively stable performance trends as they evaluate on a more stable metric: the pairwise Pearson product-moment correlation between forward features of candidate models and well-trained probe models (requiring extra training on target dataset). PARC, a simpler variant of RSA achieved by replacing the probe model with ground truth labels, also exhibits stable performance.

\begin{figure}[tbp]
	\centering
        \includegraphics[width=1\textwidth]{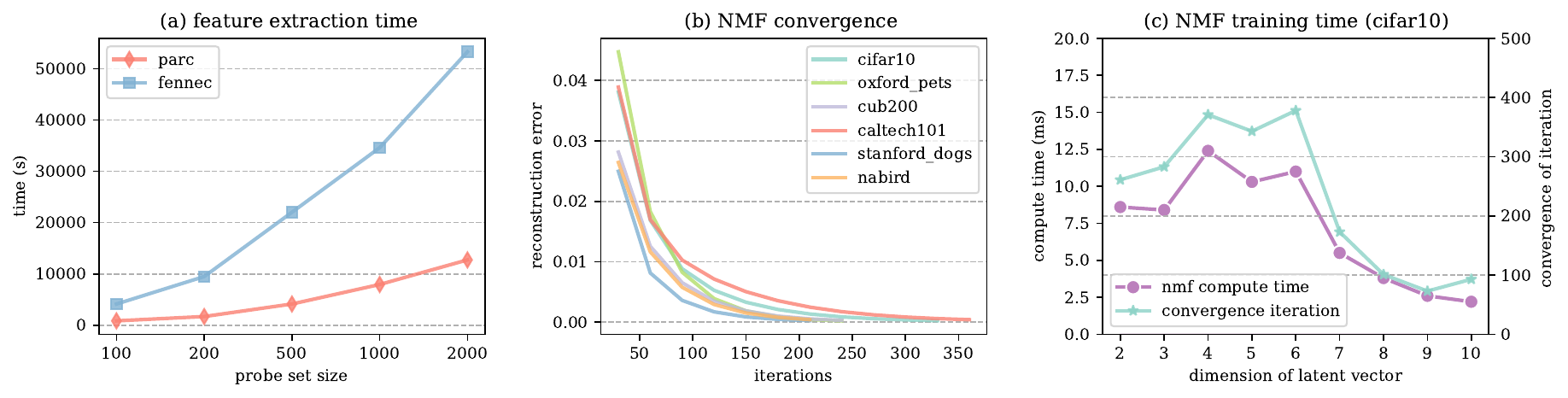}
	\caption{Analysis of feature extraction time and NMF training time. The results indicate that NMF converges rapidly, typically within 500 iterations.}
	\label{fig:mf_time}
\end{figure}

Expanding the probe set size indeed yields benefits, as evident in the results (Figure \ref{fig:probeset}(b)(d)), but the associated increase in calculation time and forward feature extraction time (Figure \ref{fig:mf_time}(a)) for these methods cannot be ignored. Notably, the calculation time for some methods shows exponential growth, a trend that should be avoided. Our method, not reliant on generating forward features during online ranking, effectively sidesteps these challenges.

\begin{table*}[tbp]
 \caption{Performances with/without transfer phase scores and meta phase scores. When excluding transfer phase scores, there is no need to compute features on the proxy model. The best performance is achieved when merging the scores of both.}
  \centering
  \begin{tabular}{l<{\centering} m{3cm}<{\centering}m{2cm}<{\centering}m{2cm}<{\centering}}
    \toprule
    Method & Features extractor time\bf (s) &Calculate time\bf(ms)  & Mean PC(\%) \\
    \midrule
    PARC w/o trans  & 0  &    $1.0 \pm 0.0$   & $53.78\pm 0.59 $ \\
    PARC w/o meta   &  635.6  &    $1.2 \pm 0.1$   & $61.89\pm 0.91$ \\
    PARC   ours & 635.6 &    $  2.2 \pm 0.1 $    & $ \bf 68.12\pm0.57 $  \\
    \midrule
    Fennec w/o trans   & 0 &  $ 3.7 \pm 0.3$  & $63.70 \pm 0.19  $ \\
    Fennec w/o meta &  420.7  & $ 1.5 \pm 0.1$ & $85.46 \pm 0.37 $  \\
    Fennec ours &  420.7 & $5.1 \pm 0.1$  & $\bf 87.79 \pm 0.29$  \\
    \bottomrule
  \end{tabular}
 
  \label{tab:ab_module}
\end{table*}

\subsection{Ablation Study}

\vspace{0.5em}
\noindent\textbf{Comparison of latent embeddings. } In the transfer phase, we obtain latent vectors representing the transferability of pre-trained models and historical tasks through matrix factorization. The dimensionality of these vectors is a key factor in this phase. By varying the dimension, we observe the experimental results depicted in Figure \ref{fig:ks}(a)(c). As the dimension increases, the performance gradually improves and eventually stabilizes. The overall trend remains consistent, showcasing the robustness of our method. 

\vspace{0.5em}
\noindent\textbf{Analysis of NMF training convergence.} We visualized the NMF decomposition process. From Figure \ref{fig:mf_time}(b)(c), it can be observed that all datasets converge steadily within 500 iterations. Additionally, we illustrate the impact of changing dimensions of latent variables on convergence. As the dimension increases, the training time of NMF initially increases and then decreases. Further investigation reveals that the increase in dimension allows NMF to converge in less iterations. Overall, training times in the millisecond range are acceptable, as our matrix are much smaller than those in traditional recommendation systems.

\vspace{0.5em}
\noindent\textbf{Comparison of different large vision models.} In the merge phase, we employ the CLIP model as a proxy to infer the latent vector for the new task. 
We compare the results using different large vision models, including CLIP~\cite{radford2021learning_clip}, Open\_Clip~\cite{schuhmann2022laionb_openclip}, and Dinov2~\cite{oquab2023dinov2}, as shown in Figure \ref{fig:ks}(b)(d).
The performance of Open\_Clip and Dinov2 outperforms the CLIP model. However, because of the intricate architecture and the vast number of parameters (e.g., Open\_Clip with 2000 million parameters), the forward process of these large models is significantly slower than CLIP. Considering the balance between accuracy and efficiency, we choose the CLIP model for feature extraction.

\begin{table*}[tbp]
 \caption{Mean PC(\%) with/without archi2vec modules. The minor differences on PARC are due to its limited number of architectures, with only four types available.}
  \centering
  \begin{tabular}{lm{3cm}<{\centering}m{3cm}<{\centering}}
    \toprule
    Method &  PARC & Fennec \\
    \midrule
     w/o archi\_graph & $12.40 \pm 0.31$  &$42.35 \pm 0.12$\\
     w/ one\_hot & $53.78 \pm 0.59$       &$55.28 \pm 0.16$ \\
     w/ archi\_graph & $53.78 \pm 0.59$   &  $63.70 \pm 0.19$ \\

    \bottomrule
  \end{tabular}
 
  \label{tab:archi_graph}
  \vspace{-1.5em}  
\end{table*}

\vspace{0.5em}
\noindent\textbf{The effect of archi2vec.} To assess the effectiveness of architectural features, we evaluate performance using only meta scores. 
From an individual model standpoint, as shown in Figure \ref{fig:archi}, visualization using vectors generated by the archi2vec module illustrates the architectural similarities. Despite ResNet18 and ResNet10 differing in layers, they share many submodules, resulting in a high cosine similarity (0.9703). In contrast, the significantly distinct architectures of ResNet18 and AlexNet yield a lower similarity (0.4673).

Regarding transferability estimation, we compare the impact of including (w/) and excluding (w/o) architectural graph features in the meta features (Table \ref{tab:archi_graph}). The inclusion of architectural features leads to significant improvements compared to relying solely on  statistical features, which fail to adequately represent transferability. For the PARC, where models have explicit architectures, one-hot and archi2vec yield similar results due to the limited architectural types.



\section{Conclusion}
In this paper, we propose a novel efficient model ranking framework that significantly reduces the overhead of evaluating the transferability of numerous models on a new dataset. The main idea is to merge historical transfer preferences between tasks and models with a similarity score for neural model structures, retrieved by our novel archi2vec module. To verify the effectiveness of our approach, we introduce a large benchmark including 105 pre-trained models, spanning over 60 different architectures. The results on two benchmarks show that our approach outperforms existing approaches in both efficiency and precision.

\bibliographystyle{splncs04}
\bibliography{main}

\appendix
\clearpage
\setcounter{page}{1}
\allowdisplaybreaks[4]

\section{Related Work}
\label{related}
We briefly review some highly related topics, including transfer learning, taskonomy, and transferability estimation, as shown in the Table \ref{tab:existing_work}.

\vspace{0.5em}
\noindent\textbf{Transfer Learning.}
Transfer learning~\cite{zhuang2020comprehensive,tan2018survey,li2020transfer} has made remarkable progress in various forms, including domain adaptation~\cite{csurka2017domain,you2019universal,wang2018deep,peng2023came,peng2024energy} and domain generalization~\cite{li2017deeper,zhou2022domain,wang2022generalizing}. Transfer learning typically involves two key stages: pre-training and fine-tuning. A meticulously designed neural network model is first trained on data-rich standard datasets such as ImageNet~\cite{deng2009imagenet}. During the fine-tuning stage, practitioners can either retrain a new head for the target task or train the entire network using the pre-trained model as an initialization. Numerous experiments have showcased that employing transfer learning not only enhances performance but also accelerates the training process~\cite{raghu2019transfusion}.

\vspace{0.5em}
\noindent\textbf{Taskonomy.}
\label{related-taskonomy}
Taskonomy~\cite{zamir2018taskonomy} stands as a pioneering work in measuring task similarity. It introduced a fully computational method for assessing relationships between different visual tasks, constructing a taxonomic map to depict these task connections. Several following-up advancements were made to reduce the computation~\cite{song2019deep,rsa@2019,dds@2020,achille2019task2vec,Song_2020_CVPR,Song_2023_ICCV}. For example, DEPARA~\cite{song2019deep} approximates Taskonomy by evaluating the similarity of attribution maps from  models.
Task2vec~\cite{achille2019task2vec} proposed estimating the Fisher information matrix linked to trained probe network parameters as the task embedding. RSA~\cite{rsa@2019} and DDS~\cite{dds@2020} uesd representation similarity analysis and duality diagram similarity analysis to measure task similarity. Despite their striking performance, all these methods required training a task-specific network anew, which incurs expensive computational costs.

\vspace{0.5em} 
\noindent\textbf{Transferability Estimation.}\label{related-core}
Of particular relevance to our research is the domain of transferability estimation. Recently, a series of more computation-efficient methods have emerged in this domain. NCE~\cite{tran2019transferability_nce} set a significant precedent by calculating the conditional entropy between two variables (label assignments of source and target tasks) as the transferability score. LEEP and NLEEP~\cite{nguyen2020leep_leep, li2021ranking_nleep} introduced the log expected empirical prediction score to estimate transferability. Bayesian LS-SVM~\cite{kim2016learning} and LogME~\cite{you2021logme} estimated the maximum value of label evidence based on forward features as the transferability score. H-score~\cite{bao2019information_hscore} utilized concepts from Euclidean information geometry to deduce the optimal log-loss of forward features and labels. TransRate~\cite{huang2022frustratingly_transrate} employed mutual information between forward features and labels to measure correlation. PARC~\cite{bolya2021scalable_parc} stood as the pioneering benchmark in this field, systematically evaluating existing methods under various model architectures. GBC~\cite{pandy2022transferability_gbc} and SFDA~\cite{shao2022not_sfda} were based on the key idea that models producing better class-wise separable forward features have superior transferability.
Etran~\cite{gholami2023etran} proposed an energy-based evaluation method, which first considered the unsupervised evaluation scenario. NCTI~\cite{wang2023far_ncti} introduced an evaluation method based on three phenomena in neural collapse. PED~\cite{li2023exploring_PED} uses latent potential energy to simulate the dynamic changes of models during the transfer process. Our method is not immersed in subtle evaluation metrics, focusing instead on an efficient model ranking method amidst a plethora of candidate models.

 Model Spider(MS)~\cite{zhang2023model_spider} is most closely related to ours, where both aim to learn vectors (tokens) denoting the transferability of models and tasks. However, our work differs from theirs in the following four aspects: \textbf{1).} MS requires introducing \textbf{an additional meta-set} for neural networks to learn the vectors of models and tasks. In contrast, inspired by recommendation systems, Fennec relies solely on the \textbf{classical collaborative filtering} method - non-negative matrix factorization to obtain vectors for models and tasks. It has fast training speed and strict convergence guarantees; \textbf{2).} The model-task similarity in MS is trained using neural networks, while we adopt the efficient inner product (additional training in the merge phase is only for obtaining vectors of new tasks); \textbf{3).} To address model ranking in unsupervised scenarios, we propose using the proxy model to obtain task vectors. However, MS must rely on labels to deduce task vectors, which is costly in practical scenarios; 
\textbf{4).} We explicitly introduce the inherent inductive bias of models into the transfer estimation and propose archi2vec to automatically learn the complex structures of models.

\begin{table*}[tbp]
	\caption{Summary of existing works. Our method does not require access to the model source, fine-tuning on the target task, or the need for forward features and labels.}
	\centering
	\begin{tabular}{l<{\centering}m{2cm}<{\centering}m{2cm}<{\centering}m{3cm}<{\centering}m{2cm}<{\centering}m{1cm}<{\centering}}
		\toprule
		Methods   & No need to access source    &  No training on the target &  No forward features required 
            & No labels required & Model space\\
		\midrule
		Taskonomy & \ding{56}  &   \ding{56} & \ding{56}  &\ding{56} &  26 \\
		Task2vec & \ding{56}  &   \ding{56} & \ding{56}  &\ding{56} &  - \\
		  DEPARA   & \ding{52}  &   \ding{56} & \ding{56}  &\ding{56} &  20 \\
            RSA      & \ding{52}  &   \ding{56} & \ding{56}  &\ding{56} &  20 \\
            DDS      & \ding{52}  &   \ding{56} & \ding{56}  &\ding{56} &  17 \\
            \midrule
            NCE      & \ding{56}  &   \ding{52} & \ding{52}  &\ding{56} &  - \\
            LEEP     & \ding{52}  &   \ding{52} & \ding{52}  &\ding{56} &  - \\
            H-Score  & \ding{52}  &   \ding{52} & \ding{56}  &\ding{56} &  8 \\
            LogME    & \ding{52}  &   \ding{52} & \ding{56}  &\ding{56} &  10 \\
            TransRate & \ding{52}  &   \ding{52} & \ding{56}  &\ding{56} & 7  \\
            PARC     & \ding{52}  &   \ding{52} & \ding{56}  &\ding{56} &  32 \\
            GBC      & \ding{52}  &   \ding{52} & \ding{56}  &\ding{56} &  9 \\
            SFDA     & \ding{52}  &   \ding{52} & \ding{56}  &\ding{56} &  11 \\
            ETran     & \ding{52}  &   \ding{52} & \ding{56}  &\ding{52} &  11 \\
            NCTI     & \ding{52}  &   \ding{52} & \ding{56}  &\ding{56} &  11 \\
            PED     & \ding{52}  &   \ding{52} & \ding{56}  &\ding{56} &  12 \\
            Model Spider     & \ding{52}  &   \ding{52} & \ding{52}  &\ding{56} &  10 \\
            \midrule
            Ours     & \ding{52}  &   \ding{52} & \ding{52}  &\ding{52} &  105 \\
            
		\bottomrule
	\end{tabular}
	\label{tab:existing_work}
\end{table*}

\section{Fennec benchmark}\label{training}
In this section, we provide a detailed construction of the Fennec benchmark.

For all tasks we use in Fennec, as shown in Table \ref{tab:expri_setting}, we use the provided training, validation, and test sets for data partitioning whenever available. If not provided, we adhere to a \textbf{6:2:2} ratio for dataset splitting—60\% of the data for training set, 20\% for constructing a validation set for hyper-parameter tuning, and the remaining 20\% for test set.

To evaluate the correlation between the estimation scores and real transfer accuracy, We meticulously fine-tune all models $\{\phi_i = (\theta_i \circ h_i)\}_{i=1}^M$ on the tasks to obtain the real transfer performances. We conduct grid search for hyper-parameter tuning over two crucial hyperparameters~\cite{you2021logme}: learning rate ( [0.1, 0.01, 0.001, 0.0001]) and weight decay ([0.01, 0.001, 0.0001, 0.00001, 0.000001]). The training is performed using the SGD for 100 epochs with full fine-tuning (no weight frozen), and early stopping is employed to expedite the training process, with a batch size of 64. The final transfer results $\{a_i^t \in {a^t}\}_{i=1}^M$ are assessed based on the performances of the fine-tuned models $(\theta_i^* \circ h_i^*)$ on the test set.

\begin{table*}[tbp]
\caption{Basic settings of the Fennec benchmark. It encompasses a variety of tasks and 105 pre-trained models. Due to space constraints, we only listed a subset of models.}
  \centering
  \resizebox{0.7\textwidth}{!}{
  \begin{tabular}{llccc}
    \toprule
      \bf Datasets & \bf Domain & \bf \# Samples &  \bf \# Classes  \\
    \cline{1-4}
  
     Sport\cite{sport} & Gesture& 14581 & 100  \\
     Yoga\cite{yoga} & Gesture & 1171 & 6  \\
     Flower\cite{nilsback2008automated_flower} & Plant & 8189 & 102  \\
     Intel\cite{intel} & Scene & 17034 & 6 \\
     Weather\cite{Weather} & Climate & 6862 & 11 \\
     Germ\cite{waquar_2022_germ} & Medical & 789 & 8  \\
     Nature\cite{roy2018effects_nature} & Scene &  6899 & 8  \\
    \midrule
    \bf Models  & \bf Venue & \bf \# Parameters & \bf \# Variants  \\
    \midrule
    AlexNet\cite{NIPS2012_alexnet} & NIPS, 2012 & 62,378,344 & 1  \\
    ZfNet\cite{zeiler2014visualizing_zfnet}   & ECCV, 2014 & 62,357,608  & 1      \\
    Vgg19\cite{simonyan2014very_vgg}   & ICLR, 2015 & 143,667,240      & 1        \\
    ResNet\cite{he2016resnet}  & CVPR, 2016 & 11,689,512     & 10        \\
    PreresNet\cite{he2016identity_preresnet}   & ECCV, 2016 & 11,687,848    & 10 \\
    ResNeXt\cite{xie2017aggregated_resnext}  & CVPR, 2017 & 7,127,336     & 3       \\
    SeResNet\cite{hu2018squeeze_senet}    & CVPR, 2018 & 28,088,024     & 7  \\
    SENet\cite{hu2018squeeze_senet}  & CVPR, 2018 & 31,366,168       & 2\\
    AirNet\cite{8510896_airnet}  & TNNLS, 2019 & 	27,425,864       & 2       \\          
    ScNet\cite{liu2020improving_scnet}  & CVPR, 2020 & 25,564,584       & 2   \\
    RegNetX\cite{radosavovic2020designing_regnetx} & CVPR, 2020 & 2,684,792       & 8      \\
    DiracNet18V2\cite{Zagoruyko2017diracnets} & ARXIV, 2017 & 11,511,784       & 2        \\
    DenseNet121\cite{huang2017densely_densenet}  & CVPR, 2017 & 7,978,856         & 4       \\
    HRNet\_V1\cite{sun2019deep_hrnet} & CVPR, 2019 & 13,187,464       & 2             \\
    ShakeShakeResNet20\cite{Gastaldi17ShakeShake}  & ICLR, 2017 & 541,082                     & 3      \\
    ShuffleNet\cite{zhang2018shufflenet} & CVPR, 2018 & 1,531,936    & 2    \\
    MobileNet\cite{howard2017mobilenets} & ARXIV, 2017 & 4,231,976      & 4 \\
    Igcv3\cite{sun2018igcv3} & BMVC, 2018 & 3,491,688      & 2  \\
    InceptionV3\cite{szegedy2016inception} & CVPR, 2016 & 23,834,568       & 4     \\
    EfficientNet\_b0\cite{tan2019efficientnet} & 	ICML 2019 & 5,288,548      & 4        \\
    \bottomrule
  \end{tabular} }
  \label{tab:expri_setting}
  \vspace{-1em}  
\end{table*}

\section{FDA scores}\label{sec:fda_deduce}
\begin{figure}[tbp]
	\centering
        \includegraphics[width=1\textwidth]{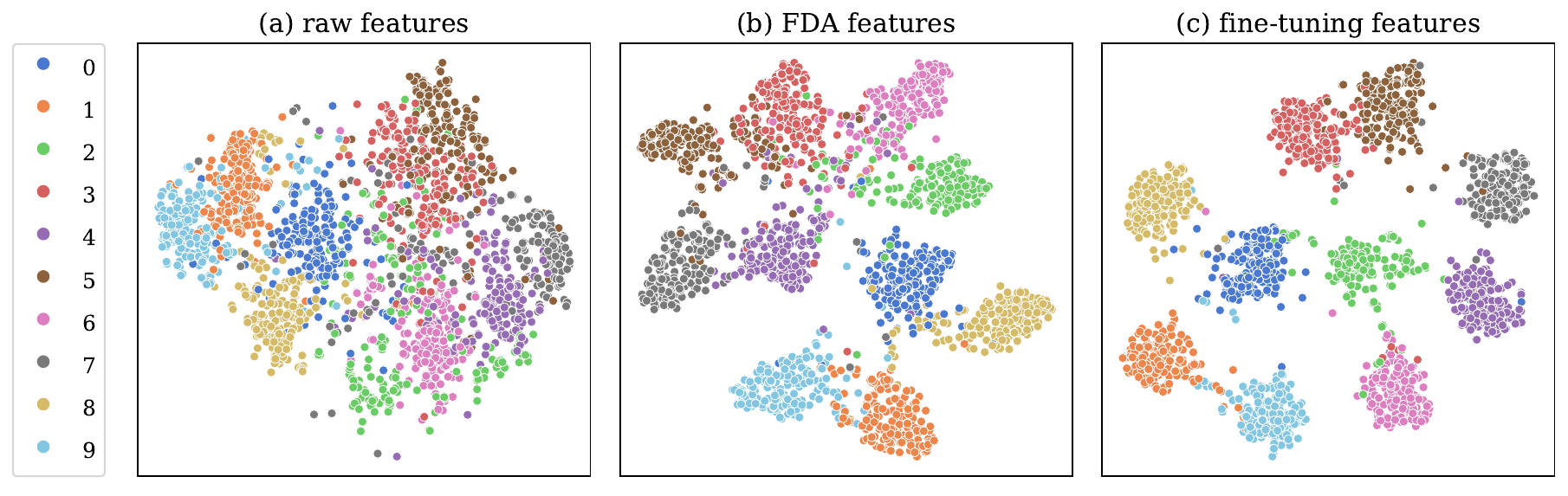}
	\caption{The (a) shows the static features generated by the model (resnet50) on the cifar10 dataset; (c) shows the features of the same model after the fine-tuning; and (b) shows the results after FDA transformation, a good simulation to the fine-tuning one.}
	\label{fig:fda}
 \vspace{-1em}  
\end{figure}

The crux of our paper lies in constructing a historical performance matrix. However, obtaining true transfer results for each model-task pair (which requires fine-tuning training) is a time-consuming process. Therefore, we use Fisher Discriminant Analysis (FDA) as a substitute for real results. We briefly outline the derivation of FDA scores. For a more detailed procedure, please refer to~\cite{mika1999fisher, ghojogh2019fisher}.

Fisher Discriminant Analysis aims to map the original data into a separable subspace where data from different categories exhibit enhanced separability, as shown in Figure \ref{fig:fda}. Fundamentally, it seeks to: (1) maximize the distance between the centers of different classes, aiming for maximal separation between data of each class; (2) minimize the scatter matrix, encouraging data within classes to be as compact as possible. Its optimization function is:
\begin{equation}
    \mathop{\rm max}\limits_{U} J(U) = \frac{U^TS_BU}{U^TS_WU} .
\end{equation}
where $S_B$ is the between-scatter of classes and $S_W$ is the within-scatter of classes.
By taking derivative, we can obtain:
\begin{equation}
    S_B u = \lambda S_W u .
    \label{eq:U}
\end{equation}
where $u$ is the column of the $U$. We can solve the Equation\eqref{eq:U} by solving generalized eigenvalues problem~\cite{ghojogh2019fisher}.
We can map the raw features using the learned matrix $U$: $\Tilde{f_{ia}}:=U^Tf_{ia}$. We then use the transformed features to calculate the FDA scores. We assume that each class $c$ follows a Gaussian distribution. For clarity, we represent $\Tilde{f_{ia}}$ as $\Tilde{f}$, omitting the subscripts.

\begin{equation}
g_c(\Tilde{f})=\dfrac{1}{(2\pi)^{d/2}|\Sigma_c|^{1/2}} 
e^{-\frac{1}{2}(\Tilde{f}-\mu_c)^T\Sigma_{c}^{-1}(\Tilde{f}-\mu_c)} .
\end{equation}
where $d$ is the dimension, $\mu_c$ represents the transformed mean value and $\Sigma_{c}$ is the covariance matrix. 
For simplicity, we set $\Sigma_{c}=\Sigma=I$(the works~\cite{shao2022not_sfda, gholami2023etran} also did same simplify). 
Next, we can obtain the final classifier based on Bayes rule: 

\begin{align}
\quad \quad    G(\Tilde{f}) & = \text{arg } \underset{c}{\text{max }}g_c(\Tilde{f})\pi_c \\
& = \text{arg } \underset{c}{\text{max }} \text{ log}(g_c(\Tilde{f})\pi_c) \\
& = \text{arg } \underset{c}{\text{max }}\lbrack - \text{log}((2\pi)^{d/2}) -  \nonumber \\
& \quad   \quad   \frac{1}{2}(\Tilde{f}-\mu_c)^T(\Tilde{f}-\mu_c)+\text{log}(\pi_c)  \rbrack \\
& = \text{arg } \underset{c}{\text{max}}\lbrack-\frac{1}{2}(\Tilde{f}-\mu_c)^T(\Tilde{f}-\mu_c)+\text{log}(\pi_c)  \rbrack \\
& = \text{arg } \underset{c}{\text{max}} \Tilde{f}^T\mu_c-\frac{1}{2}\mu_{c}^{T}\mu_c-\frac{1}{2}\Tilde{f}^T\Tilde{f} +\text{log}(\pi_c) \\
& = \text{arg } \underset{c}{\text{max }} \Tilde{f}^T\mu_c-\frac{1}{2}\mu_{c}^{T}\mu_c +\text{log}(\pi_c) .
\end{align}
where $\pi_c=\frac{n_c}{n_j}$ is the prior probability. we remove  $-\frac{1}{2}\Tilde{f}^T\Tilde{f}$  because this term remains constant for the computation of each sample. Correspondingly, the formula for the score of an individual class $c$ can be obtained by eliminating the $\text{arg } {\text{max}}$ term. Thus, we derive the form of Equation\eqref{eq:fda_final}.

\begin{table*}[tbp]
\caption{Names of nodes in archi2vec. archi2vec can automatically identify key nodes and edges in the architecture of a arbitrary neural model.}
    \centering
    \resizebox{\textwidth}{!}{ 
    \begin{tabular}{lllll}
    \toprule
    \multicolumn{5}{c}{Nodes in archi2vec}  \\
    \midrule
       AdaptiveAvgPool2DBackward0  & AdaptiveMaxPool2DBackward0 & AddBackward0 & AddmmBackward0 & AvgPool2DBackward0  \\ 
       AvgPool3DBackward0  & CatBackward0 & CloneBackward0 & ConstantPadNdBackward0 &  ConvolutionBackward0    \\
        DivBackward0 & ExpandBackward0 & HardtanhBackward0 & IndexSelectBackward0 & MaxBackward0      \\
     MaxPool2DWithIndicesBackward0 & MeanBackward1 & MulBackward0 & NativeBatchNormBackward0 & Variable \\
       PowBackward0 & PreluBackward0 & ReluBackward0 & RepeatBackward0 & ReshapeAliasBackward0 \\
       SigmoidBackward0 & SliceBackward0 &  SoftmaxBackward0  & SplitWithSizesBackward0  & SqueezeBackward1 \\
       SumBackward1 & TBackward0 & TransposeBackward0 & UnsqueezeBackward0 & UpsampleBilinear2DBackward1 \\
        UpsampleNearest2DBackward1 & ViewBackward0 \\
    \bottomrule
    \end{tabular} }
    
    \label{tab:atoms}
\end{table*}

\section{Additional Results}

\begin{figure*}[tbp]
	\centering
        \includegraphics[width=0.95\textwidth]{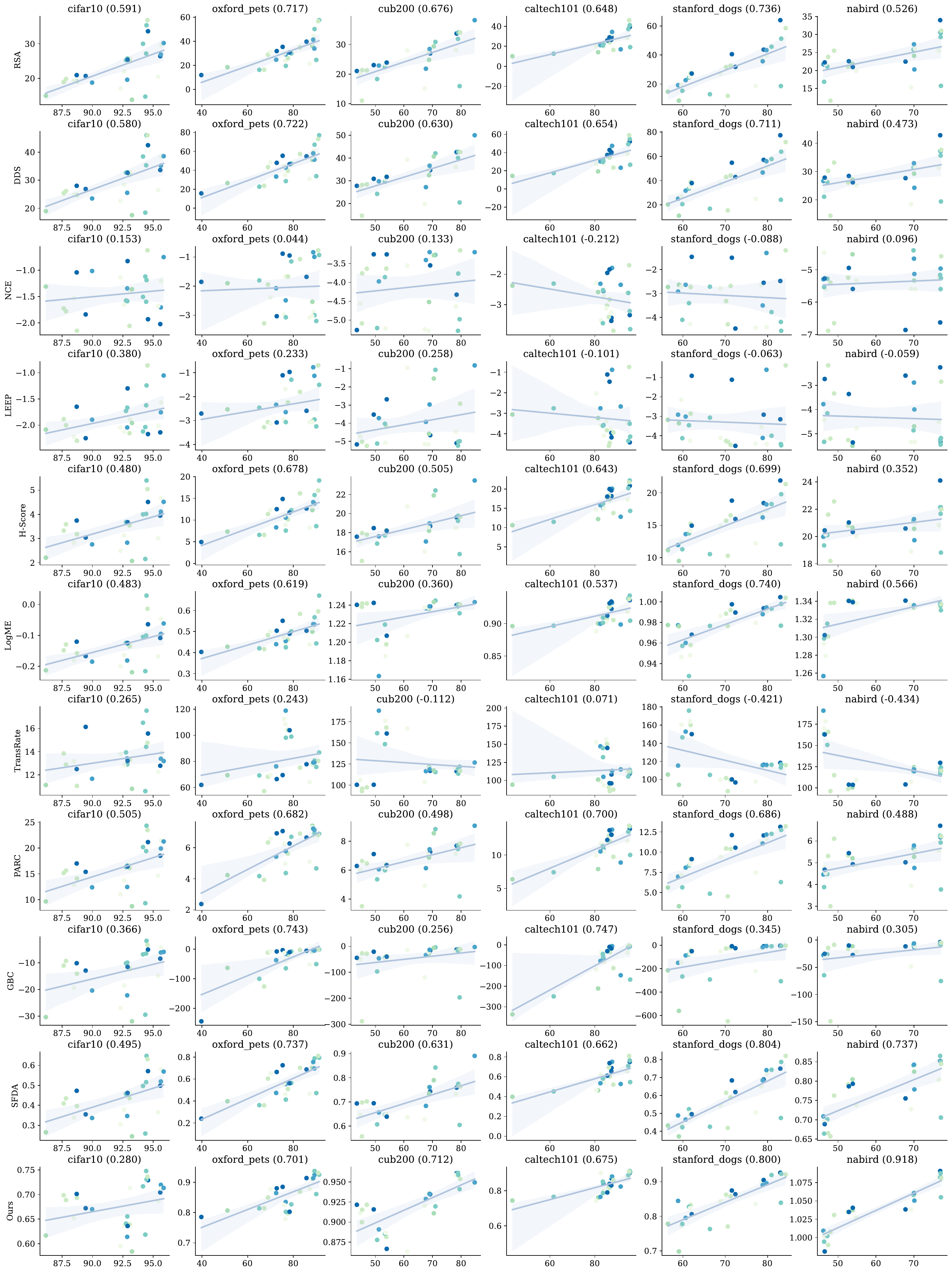}
	\caption{The detailed results on the PARC benchmark. The horizontal axis denotes the actual transfer results, and the vertical axis denotes the evaluation results of each method. The title includes the Pearson correlation coefficient.}
	\label{fig:parc_sep}
\end{figure*}

\vspace{0.5em}
\noindent\textbf{PARC.}
In Table \ref{tab:main_result1}, we show the average results of existing methods on the PARC benchmark. These results are computed by taking the averages of each method across all target tasks. In this section, we show the detailed results of each method on individual datasets, as shown in Figure \ref{fig:parc_sep}.

\begin{figure*}[tbp]
	\centering
        \includegraphics[width=0.95\textwidth]{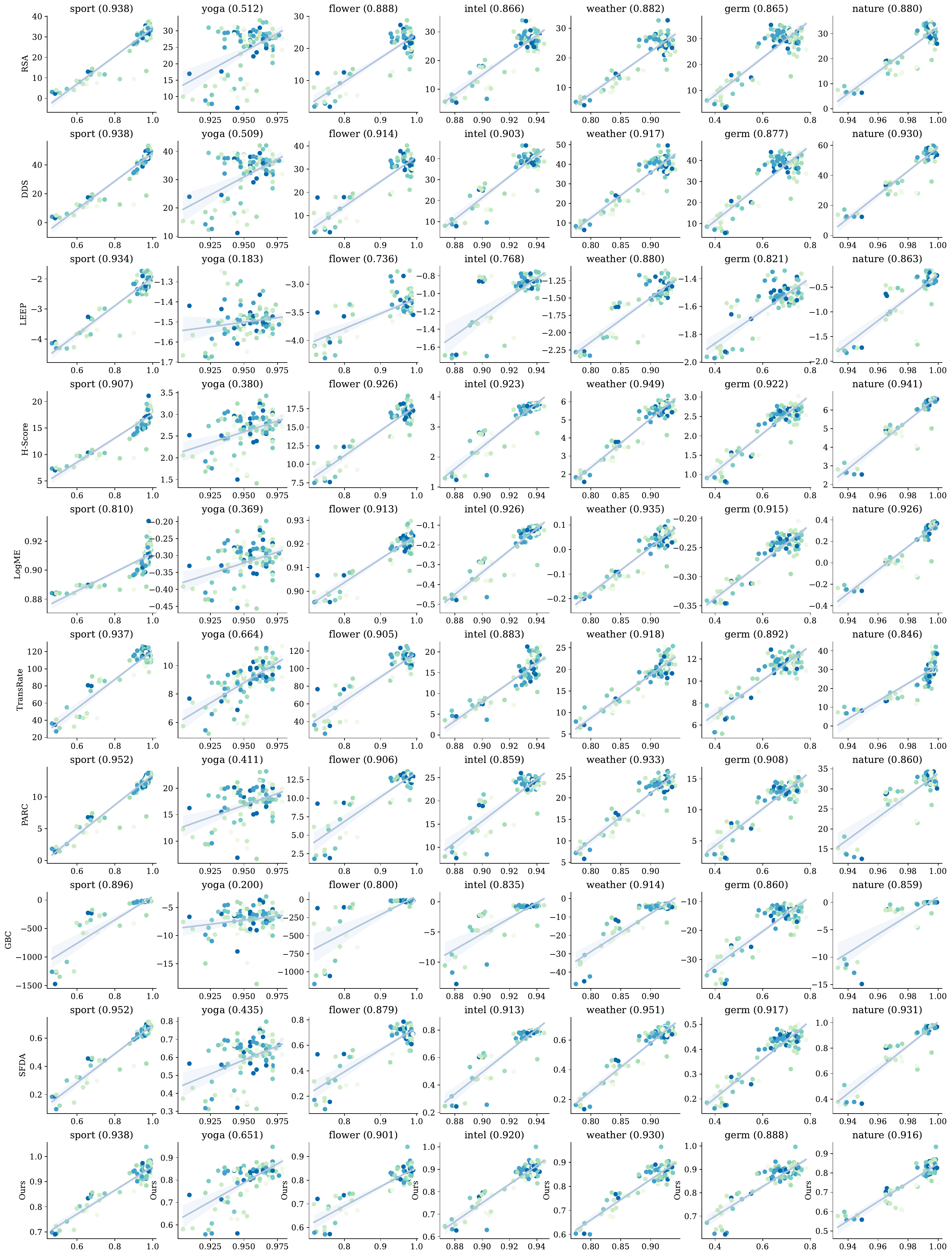}
	\caption{The detailed results on the Fennec benchmark. The horizontal axis denotes the actual transfer results, and the vertical axis denotes the evaluation results of each method. The title includes the Pearson correlation coefficient.}
	\label{fig:fennec_sep}
\end{figure*}

\vspace{0.5em}
\noindent\textbf{Fennec.}
On the Fennec benchmark, the detailed Pearson correlation coefficients for each method across various target tasks are shown in Figure \ref{fig:fennec_sep}.

\begin{figure*}[tbp]
	\centering
        \includegraphics[width=0.7\textwidth]{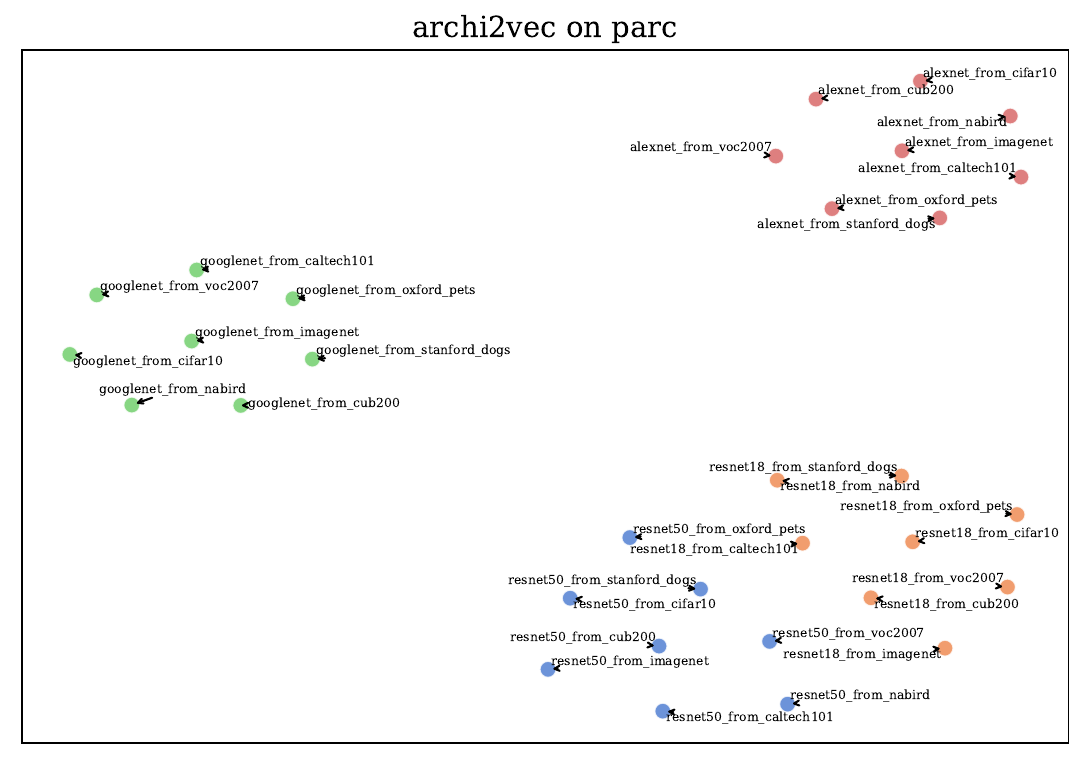}
	\caption{The T-SNE visualization of the model vectors on PARC benchmark generated by archi2vec. The models in the ResNet family are closely related.}
	\label{fig:archi_vis_parc}
\end{figure*}

\begin{figure*}[tbp]
	\centering
        \includegraphics[width=0.98\textwidth]{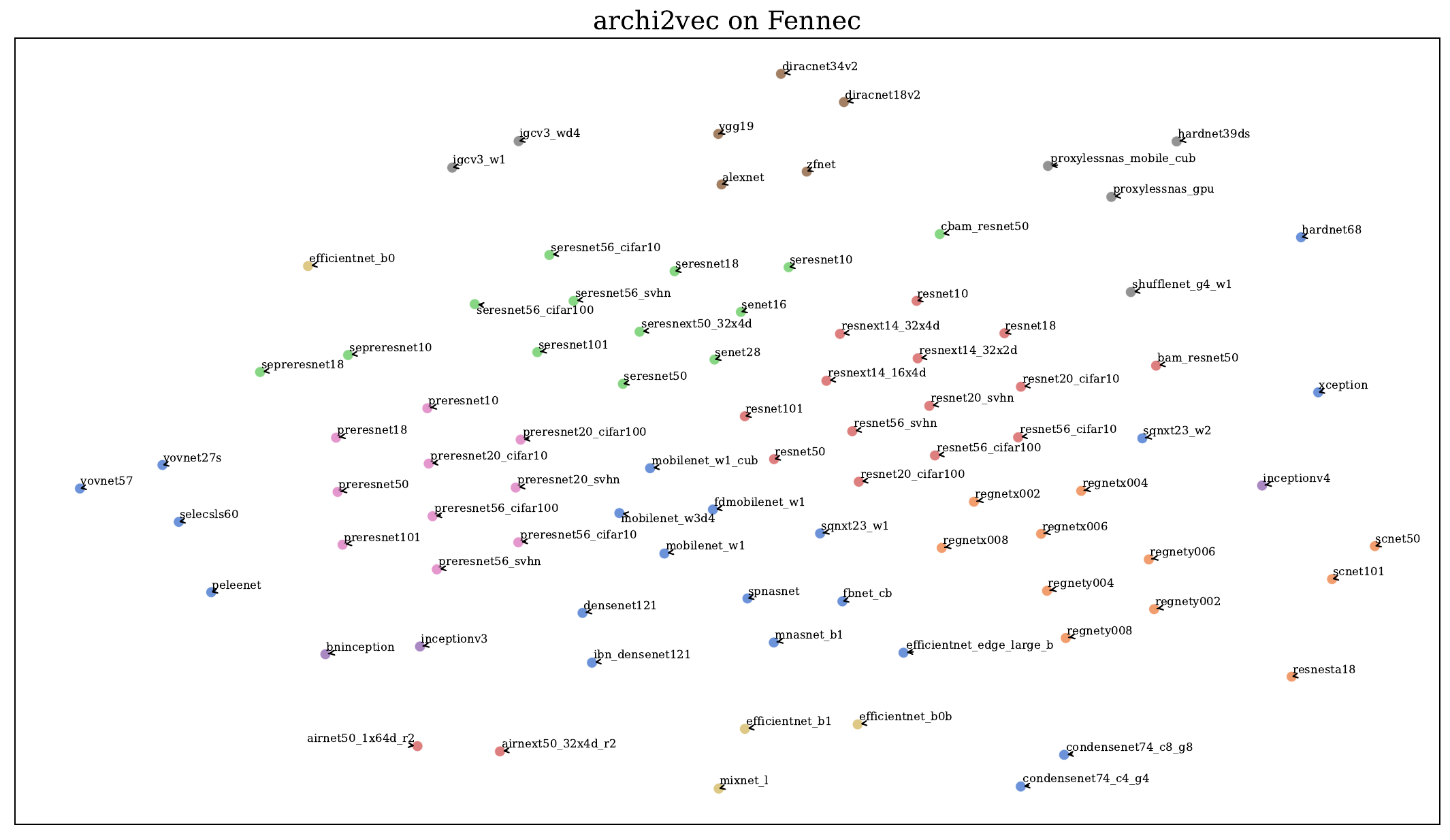}
	\caption{The visualization of the Fennec model vectors generated by archi2vec using T-SNE. Archi2vec method can automatically identify models with similar architectures. }
	\label{fig:archi_vis_fennec}
\end{figure*}

\vspace{0.5em}
\noindent\textbf{archi2vec.}
In Section \ref{sec:meta_phase}, we present the archi2vec method, which has the capability to automatically encode complicated neural network architectures into vectors. The atomic nodes of the graphs are composed of 37 types: $(A_i)_{i=1}^M \in (atom_i)_{i=1}^{37}$, as listed in Table \ref{tab:atoms}. The names of atomic nodes correspond to type names defined in the PyTorch framework. Furthermore, we visualize the model vectors generated by archi2vec, as shown in the Figure \ref{fig:archi_vis_parc},\ref{fig:archi_vis_fennec}. 
From the visualization results on PARC, it is evident that our method effectively represents the architectures of the models. Due to the limited diversity in structures on this benchmark, our method successfully clusters similar architectures together, bringing ResNet18 and ResNet50 closer, for instance. As for the Fennec benchmark, which encompasses a more diverse set of model architectures, we only showcase the visualization results for a subset of models. Our method adeptly clusters models from various series, such as ResNet, PreResNet, and EfficientNet. Notably, this process requires no manual intervention and expert knowledge.

\begin{figure*}[tbp]
	\centering
        \includegraphics[width=0.9\textwidth]{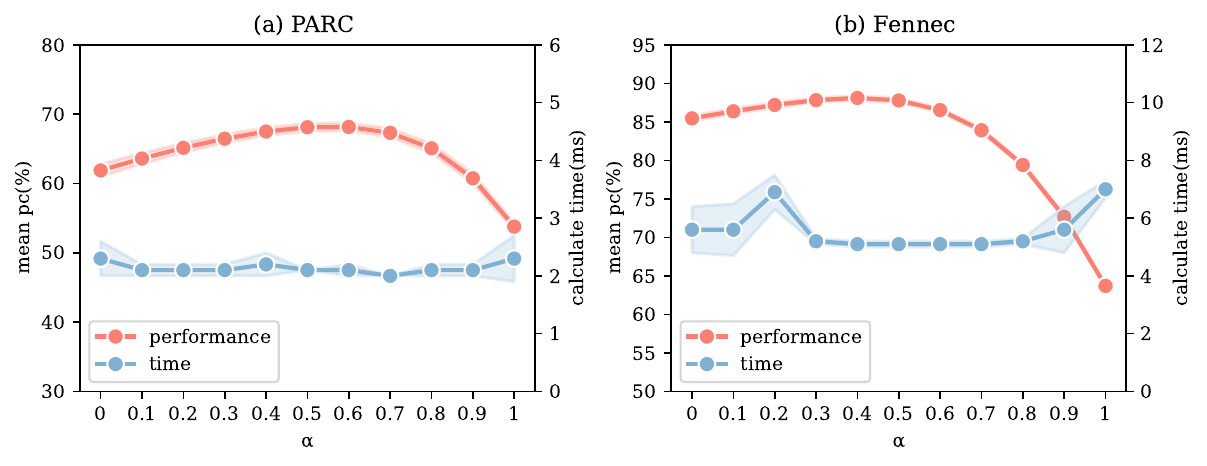}
	\caption{Performance on different weight $\alpha$. It is used to adjust the weights between the transfer and meta stages. The default value in this work is 0.5.}
	\label{fig:alpha}
\end{figure*}

\vspace{0.5em}
\noindent\textbf{Comparison of $\alpha$.} We merge the scores of the transfer phase and meta phase as the final estimation scores. The impact of these two phases is controlled by $\alpha$. We vary the value of $\alpha$ from 0 to 1 to observe the changes in the results. As $\alpha$ increases, the proportion of the meta phase gradually increases, while the proportion of the transfer phase gradually decreases.
As shown in Figure \ref{fig:alpha}, as we gradually integrate the features from the meta phase, the performance steadily improves to its peak (the weight associated with the highest point on both PARC and Fennec benchmarks is 0.6 and 0.4, respectively). Beyond this point, the performance begins to decline gradually due to the decreased involvement of the transfer phase. In reality, the optimal weight value $\alpha$ may vary for different tasks and models. In this paper, we use 0.5 by default. We intend to investigate this as part of future work.

\begin{table}[tbp]
 \caption{Comparison with unsupervised methods. Our method not only has less feature extraction time but also exhibits better correlation.}
  \centering
  \resizebox{\textwidth}{!}{
  \begin{tabular}{l<{\centering}m{2cm}<{\centering} m{2cm}<{\centering}m{2cm}<{\centering}m{2cm}<{\centering}  m{2cm}<{\centering}m{2cm}<{\centering}m{2cm}<{\centering} | m{2cm}<{\centering}m{3cm}<{\centering}}
    \toprule
    Method &  Sport & Yoga & Flower & Intel & Weather & Germ & Nature & Mean PC & Features extractor time(s) \\
    \midrule
     energy & $49.87\pm2.13$  & $18.32\pm 2.83$ & $41.08\pm 1.81$  &  $33.55\pm 2.55$ &  $52.23\pm 3.37$ &
     $41.48\pm 1.28 $ & $34.25\pm 3.12$ &  $38.68\pm 0.42$ & 22015.9\\
    Ours & $93.79\pm 0.22$  & $65.15\pm 0.64$ & $90.14\pm 0.56$  &  $92.04\pm 0.25$ &  $92.99\pm 0.24$ &
     $88.79\pm 0.44 $ & $91.64\pm 0.16 $ & $87.79\pm 0.29$ & 420.7 \\

    \bottomrule
  \end{tabular}}
  \label{tab:unlabeled}
\end{table}

\begin{figure*}[tbp]
	\centering
        \includegraphics[width=1\textwidth]{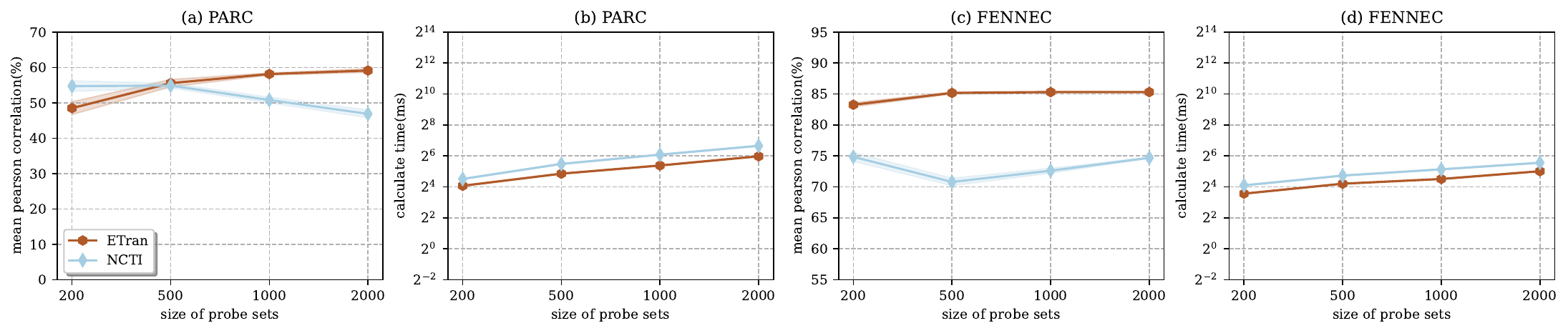}
	\caption{Comparison of performance and time of baselines varying on  probe sizes.}
	\label{fig:probeset_etran}
 \vspace{-1em}  
\end{figure*}

\vspace{0.5em}
\noindent\textbf{Size of probe sets.} In our experiments, we observe that when the number of samples in the probe set is small (e.g., 100 samples), ETran and NCTI generate partially incorrect scores (e.g., the energy term in ETran resulted in NaN, not a number). Therefore, we vary the number of samples in the probe set from 200 to 2000 to observe the impact of the probe set size on the results. 
From Figure \ref{fig:probeset_etran}, it can be observed that as the size of the probe set increases, the accuracy of ETran gradually improves, while NCTI shows a declining trend. It is also evident that the computational time for both methods increases significantly. Both of these methods require the involvement of forward features for scoring, and the time taken for feature extraction cannot be overlooked, as indicated in Table \ref{tab:main_result1}.

\vspace{0.5em}
\noindent\textbf{Comparison of the unsupervised version of ETran.} In ETran~\cite{gholami2023etran}, solely relying on the energy generated by forward features enables handling of the unlabeled scenario. To our knowledge, this is the only method capable of ranking in an unsupervised setting. Nevertheless, it is important to note that it computes the energy score of forward features, so the extraction of forward features is indispensable. This can be observed from Table \ref{tab:unlabeled}. Compare to ETran, which only uses energy scores of forward features, our method has less features extraction time and better correlation in unsupervised scenarios.

\end{document}